\documentclass[onefignum,onetabnum, 11pt]{siamart251216}

\usepackage{lipsum}
\usepackage{amsfonts}
\usepackage{amsmath,mathtools,enumerate}
\usepackage[numbers]{natbib}
\usepackage{graphicx}
\usepackage{epstopdf}
\usepackage{multirow}

\usepackage[noend]{algorithmic}
\ifpdf
  \DeclareGraphicsExtensions{.eps,.pdf,.png,.jpg}
\else
  \DeclareGraphicsExtensions{.eps}
\fi

\usepackage{array}
\newcolumntype{C}[1]{>{\centering\arraybackslash}p{#1\dimexpr1.6cm}}

\newcommand{\beq}{\begin{equation}}
\newcommand{\eeq}{\end{equation}}
\newcommand{\beqa}{\begin{eqnarray}}
\newcommand{\eeqa}{\end{eqnarray}}

\newcommand{\dd}{\mathrm{d}}
\newcommand{\sdef}{\stackrel{def}{=}}

\newcommand{\bfQ}{\boldsymbol{Q}}

\newcommand{\bfR}{\boldsymbol{R}}

\newsiamremark{remark}{Remark}
\newsiamremark{hypothesis}{Hypothesis}
\crefname{hypothesis}{Hypothesis}{Hypotheses}
\newsiamthm{claim}{Claim}

\headers{Online PaLD}{J. D. Foley and J. T. Lee}

\title{Online Partitioned Local Depth for semi-supervised applications\thanks{Submitted to the editors DATE.}}

\author{John D.  Foley\thanks{Metron, Inc., 1818 Library St., \# 600, Reston, VA (\email{foley@metsci.com}, \email{leej@metsci.com})}
\and Justin T. Lee\footnotemark[1]}

\usepackage{amsopn}

\ifpdf
\hypersetup{
  pdftitle={An online algorithm for Partitioned Local Depth for semi-supervised applications},
  pdfauthor={J. D. Foley and J. Lee}
}
\fi

\newtheorem{introtheorem}{Theorem}

\begin{document}

\title{Online Partitioned Local Depth for semi-supervised applications}

\maketitle

\begin{abstract}
We introduce an extension of the partitioned local depth (PaLD) algorithm 
that is adapted to online 
applications such as semi-supervised prediction. PaLD is best known for unsupervised, \emph{parameter-free} clustering, but its robustness is
based on triples of data points, making exact analysis computationally expensive. Research is ongoing to improve the scalability of the underlying discrete algorithm and expand the breath of PaLD's applications.

The new algorithm we present, online PaLD, is well-suited to situations where it is possible to pre-compute a cohesion network from a reference dataset.
After $O(n^3)$ steps to construct a queryable data structure, online PaLD can extend the cohesion network to a new data point in $O(n^2)$ time.
Our approach complements previous speed up approaches 
based on approximation and parallelism. 
In practical terms, online PaLD makes larger datasets accessible to exact analysis with a relatively simple implementation. 
We present applications to online anomaly detection and semi-supervised classification 
for health-care datasets as initial illustrations of online PaLD's potential to expand applications of the PaLD framework.
\end{abstract}



\section{Introduction}
This article introduces an extension of the partitioned local depth (PaLD) algorithm to compute data cohesion adapted to online applications. PaLD excels at unsupervised, parameter-free clustering (highlighted in the Proceedings of the National Academy Sciences \cite{bmm22}) and is built on a probabilistic foundation with desirable theoretical properties \cite{bfl24,bmm22,moo23}.  
The PaLD framework learns a network of communities in data that contains structural information beyond canonically induced clusters. PaLD has been applied to classification \cite{lan22}, data depth \cite{evans2024two} and network science \cite{blz25, khoury2024informative}. 
The 
basis of the framework's parameter-free robustness is relative comparisons among triples of data points, but this necessitates $O(n^3)$ comparisons to compute cohesion between all pairs of data points.  Scaling to large datasets requires approximations \cite{bddp22}.

Our extension, online PaLD, is adapted to applications where (i) a reference dataset is fixed and (ii) many individual test points may need to be compared to the reference cohesion network. After an upfront cost of making $O(n^3)$ relative comparisons, adding one more datum will entail $O(n^2)$ more relative comparisons.  This makes larger datasets accessible to exact analysis and complements previous work on scalability \cite{bddp22,dg24sequential, lan22}. 

\subsection{Related Work}
The relatively new partitioned local depth (PaLD) framework \cite{bm22,bmm22} has been developed theoretically \cite{bfl24, bmm22, moo23} and practically in terms of scalability \cite{bddp22,dg24sequential, lan22} and applications to data \cite{blz25, bmm22,  evans2024two, khoury2022partitioned, khoury2024informative}.
The basic theory is developed in the seminal work \cite{bmm22} which is extended by the generalized PaLD approach \cite{bfl24} to manage variation and uncertainty. In \cite{moo23}, the foundations are reconsidered, including analysis of probabilistic weights on data and consistency with combining weights. 
Applications of PaLD include biological \cite{bmm22, evans2024two, khoury2022partitioned, khoury2024informative} and social settings \cite{ bfl24, blz25, bmm22}. 

Two directions to develop the scalability of PaLD include approximation techniques \cite{bddp22} and computer science approaches to address how the data is processed \cite{dg24sequential, lan22}. The 
Partitioned Nearest Neighbors Local Depth (PaNNLD) algorithm \cite{bddp22} computes an approximation of cohesion by analyzing the $k$-nearest neighbor digraph. 
By employing coin flips to resolve low impact triplet comparisons, a principled approximation (with bounded error) is obtained.  
Work to improve data processing includes a simple Python implementation adapted to GPU processing \cite{lan22}
and analysis to design sequential
and shared-memory parallel algorithms for PaLD \cite{dg24sequential}.

\subsection{Summary of Contribution}
After $O(n^3)$ steps to construct a queryable data structure, online PaLD can extend the cohesion network to a new test point in $O(n^2)$ time. Specifically, the new algorithm computes the new neighborhood of the test datum in the extended cohesion network as well as an updated strong cohesion threshold in $O(n^2)$ steps. 

\begin{introtheorem}\label{thm:query}
  Given a reference dataset $S$ of size $n$, the online PaLD algorithm constructs a queryable data structure in $O(n^3)$ time such that the (strong) neighborhood of a test point $t \notin S$ is computable in $O(n^2)$ time.
\end{introtheorem}

These determinations are exact, implementation  is relatively simple, and a basic validation of the speedup is in line with complexity analysis. See \cref{tab:speedup} below for speed up examples and \Cref{sec:query} for algorithmic details, including a description of our prototype implementation in \cref{sec:speed_up}. Details for our computational complexity results are provided in \Cref{sec:derive}.

\begin{table}[htbp]
\footnotesize
\caption{Mean run times for baseline partitioned local depth (PaLD) algorithm on $n+1$ data points vs. online PaLD to compute the neighborhood of the $n+1$-st point $t$ (averaged over 100 sample runs).}\label{tab:speedup}
\begin{center}
  \begin{tabular}{|c|c|c|} \hline
   $n$ & \bf Base PaLD (sec) & \bf Online PaLD (sec)\\ \hline
    7 & 2e-3 & 1.6e-4 \\
    15 & 3e-3 & 2.3e-4 \\ 
    239 & 0.93 & 0.032 \\
    499 & 6.2 & 0.17 \\
    787 & 28.4 & 0.434 \\ 
    999 & 64.1 & 0.844 \\ 
    1999 & 612 & 4.15 \\ \hline
  \end{tabular}
\end{center}
\end{table}

Remarkably, the memory needed for online applications can be reduced to $O(n^2)$; see \cref{thm:queryfull} for a combined runtime and memory runtime result. Our results are obtained by translating analysis of how adding one test point $t$ impacts the cohesion network into an algorithm to compute the (strong) neighborhood of $t$. 

Online PaLD opens the potential for applications to online anomaly detection and semi-supervised prediction for moderately sized data sets without the need for approximation techniques \cite{bddp22} or more complex implementations \cite{dg24sequential, lan22}.
For instance, the THINGS initiative \cite{hebart2019things} has developed 1,854 (highly-curated) concepts as a bridge between research in human cognition and artificial intelligence (see also \Cref{sec:discussion} below). Based on \cref{tab:speedup}, online PaLD would be suitable to build an online tool to relate test data to THINGS concepts. 

To explore applications to anomaly detection and prediction, experimentation across datasets and algorithmic details will be needed. Since online PaLD is relatively straightforward to implement, it can readily be added to an existing experimental environment to accelerate testing   
(see \cref{sec:prediction} for related discussion).
For big data, 
the presented approach might provide a complementary  means to reduce algorithmic runtime, 
but this would require future work to hybridize with approximation approaches  \cite{bddp22}.  See \Cref{sec:discussion} for discussion of avenues for future work.

\subsection*{Organization of the paper}
Our main results and algorithm design are in \Cref{sec:query} including a basic validation of the speed up (see \cref{tab:speedup}).
Applications to health-care data appear in \Cref{sec:application}, derivation details are provided in \Cref{sec:derive}, and discussion of directions for future work and final remarks follow in
\Cref{sec:discussion}.

\section{A query formulation of PaLD}
\label{sec:query}

The novel partitioned local depth (PaLD) algorithm adapts a dissimilarity measure \cite{bmm22}. PaLD’s intrinsic basis on relative comparisons among triples of data aids its robustness but results in an algorithm with $O(n^3)$ computational steps.  However, after an upfront cost of $O(n^3)$ computational steps is paid, considering one more datum entails $O(n^2)$ further relative comparisons. For a setting in which (i) a reference dataset is fixed and (ii) many individual test points may need to be compared to reference data (see, e.g., \cite{lan22}), it would be advantageous to pay the upfront cost to analyze the reference dataset, once and for all, and then pay the marginal cost to reason about test points, as needed.  See \Cref{sec:application} for example applications to health-care datasets.  

Here we realize this idea as an explicit algorithm.    After reviewing PaLD, we present our new query algorithm and an empirical validation of speedup. 

\begin{figure}
    \centering
    \includegraphics[width=0.95\textwidth,trim={2cm 4cm 2cm 5cm},clip]{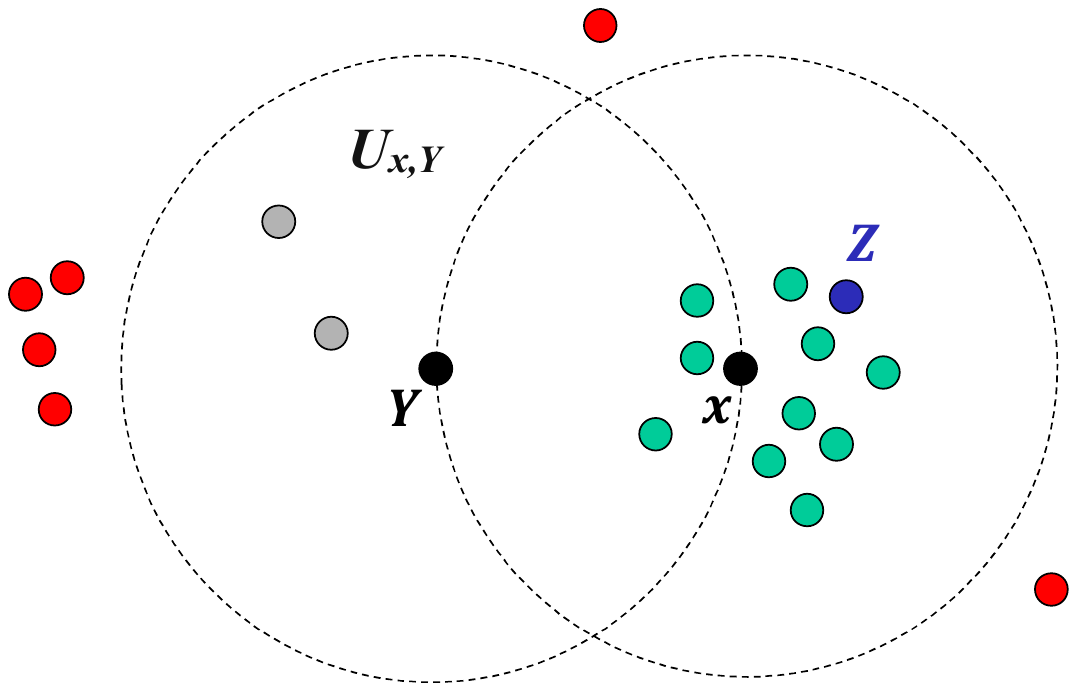}
\caption{Example local focus for a fixed point $x$ and a random point $Y$ in two dimensions. Red points are outside the focus. Those in green (and $Z$ in blue) are in the focus and closer to $x$ while those in grey are closer to $Y$. Figure from \cite{bfl24} with permission.}
\label{fig_Uxy}
\end{figure}

\subsection{Review of PaLD}
The partitioned local depth (PaLD) algorithm reasons over a finite set $S$ of data with a pairwise notion of dissimilarity or distance $\dd(x,y)$.  For any data pair $(x,y)$, the relevant local data is the set of points $z$ within $\dd(x,y)$ of $x$ or $y$:
\beq
U_{x,y} \sdef \{z \in S \mid \dd(z,x) \le \dd(y,x)  ~{\rm or }~ \dd(z,y) \le \dd(x,y)\}.
\label{local_focus}
\eeq
\noindent called the \emph{local focus} of the pair $(x, y)$; compare \cref{fig_Uxy} and note that $U_{x,y} = U_{y,x}$.  \emph{ Cohesion} is a probabilistic quantity obtained from an intuitive sampling process: select $Y \neq x$ uniformly and then a uniform $Z \in U_{x, Y}$. The cohesion of $w$ to $x$ is: 
\beq\label{eq:cohesion}
c_{x,w}\sdef P\left( Z=w,\dd(Z,x)<\dd(Z,Y) \right) + \frac{1}{2} P\left( Z=w,\dd(Z,x)=\dd(Z,Y) \right).  
\eeq	 	
This sampling process is illustrated in \cref{fig_Uxy}: once $Y \neq x$ is selected (black points), a uniform 
$Z \in U_{x, Y}$ (blue) is selected. In \cref{fig_Uxy}, red points are outside the local focus and thus, not sampled. Here the blue point $w$, any green point, and $x$ itself are potential elements $Z$ of $U_{x, Y}$ such that $\dd(Z, x) < \dd(Z, Y)$ but only $Z = w$ contributes to the cohesion of $w$ to $x$: $c_{x,w}$.  For additional discussion
and applications, including the relationship to data depth as well as considerations for
parameter-free clustering and near-neighbors, see \cite{bmm22}.  Theoretical properties of the framework are covered in \cite{bfl24,bmm22,moo23}. 

\subsubsection*{Computing cohesion}
As various 
$Y \neq x$ are sampled, further contributions to the cohesion of $w$ to $x$ accumulate. Explicitly, if  $|S| = n$, then: 
\beq
\label{eq:compute_cohension}
c_{x,w}= \frac{1}{n-1} \sum_{y \in S, y \neq x} \frac{I(w \in U_{x,y})}{|U_{x, y}|}I(w,x,y)
\eeq	
for:
\beqa
\label{IPaLD}
I(w,x,y) &=& 
\begin{cases}
1, & \text{if } \dd(w, x) < \dd(w, y) \\
1/2, & \text{if } \dd(w, x) = \dd(w, y)\\
0, & \text{otherwise}
\end{cases} ~~. 
\eeqa
\noindent 
where $1 / (n-1)$ is the sampling weight for $Y \neq x$, $1 / |U_{x, y}|$ is the sampling weight for  $Z \in U_{x, y}$  and the indicator function $I(w, x, y)$ divides support, as determined by the definition of the cohesion of $w$ to $x$ \cref{eq:cohesion}. 
In the more general settings (e.g., when quantifying data uncertainty), 
the sampling weight for $Z \in U_{x, y}$ can be abstracted to the concept of \emph{local relevance} and the indicator is captured by \emph{support division} within the generalized PaLD framework; see \cite{bfl24} for further discussion and more general notation.

\begin{algorithm}
\caption{PaLD: reference implementation to compute cohesion \cref{eq:cohesion}}
\label{alg:paldref}
\begin{algorithmic}
\STATE{Given $D(S,S)=\{ \dd(x,y)\}_{x,y=1}^n$, initialize  $C \leftarrow \{ 0\}_{x,y=1}^n$.}
\FOR{$x = 1$ to $n$}
\FOR{$y = 1$ to $n$ such that $y \neq x$}
\STATE{$U_{x, y} \leftarrow \emptyset$}
\FOR{$w = 1$ to $n$}
\IF{$\dd(w,x) \le \dd(y,x)$ or $\dd(w,y) \le \dd(x,y)$}
\STATE{$U_{x, y} \leftarrow U_{x, y} \cup \{ w\} $}
\ENDIF
\ENDFOR
\FOR{$w \in U_{x, y}$}
\STATE{$C_{x,w} \leftarrow C_{x,w} + \frac{I(w,x,y)}{|U_{x, y}|} $}
\ENDFOR
\ENDFOR
\ENDFOR
\STATE{$C \leftarrow C / (n-1)$}
\RETURN $C$
\end{algorithmic}
\end{algorithm}

Based on the above, a reference implementation for PaLD to compute a cohesion matrix from $O(n^3)$ relative comparisons is given as \cref{alg:paldref}. This approach is only lightly optimized in PaLD's open-source R package \cite{dmb25} (e.g., by applying vectorization strategies).  See \cite{bddp22} for approximation techniques and \cite{dg24sequential,lan22} for examples of parallel processing approaches. 

The \emph{cohesion network} $G_S$ is the weighted, undirected graph with node set $S$ and edge weights:
\beq
w_{x,y} \sdef \min(c_{x,y}, c_{y,x}).  
\label{sym_cohesion}
\eeq
\noindent
One can consider these $w_{x,y}$ as being a locally adapted (and probabilistically normalized) version of $\dd(x, y)$\footnote{The self-cohesion values $c_{x,x}$  can be interpreted as a localized measure of depth in the sense that the explicit formula \cref{eq:compute_cohension} indicates that large self-cohesion corresponds to membership in relatively small local foci $U_{x, y}$, due to being relatively similar to local data.}: if $w_{x,y}$ is higher, then  the pair $(x, y)$ is more closely tied, in a locally adapted, relative sense. 

Under the standard, mild assumption that $\dd(z, z) < \dd(z, y)$ for all distinct $z, y \in S$, there is a natural probabilistic threshold \cite{bmm22}:
\beq
\tau = \frac{1}{2n} \sum_{x\in S}c_{x,x} \label{threshold}
\eeq
\noindent	 	
that provides a canonical and useful point of reference for these probabilistically derived quantities; see \cite[Eqn.\ 4]{bmm22} for the general definition and further discussion. A link in $G_S$ is strong if and only if $w_{x,y} \geq \tau(S, \dd)$\footnote{Such strong links form a subnetwork of $G_S$ called the cluster network $G_S^*$, whose components determine a data-driven partitioning of $S$ into clusters without tuning paramters. Strong performance relative to clustering benchmarks was shown in \cite{bmm22}. }. Thus, by locally adapting dissimilarity, a natural threshold straightforwardly determines which $y$ are “near” $x$: $y$ such that $w_{x,y} \geq \tau(S, \dd)$.  

For our motivating setting—in which $S$ is a fixed reference dataset and there is an individual test point $t$ for online analysis—if $w_{t,y} < \tau(S, \dd)$ for all $y$ in the reference set $S$, then $t$ can be considered an outlier\footnote{For comparison, when PaLD is applied to parameter-free clustering \cite{bmm22}, any point with no strong links forms its own cluster of size one.}. Likewise, $y$ with $w_{t,y} \geq \tau(S, \dd)$ are natural reference points to compare to the test point, e.g., for semi-supervised classification \cite{lan22} similar to k-nearest neighbor approaches. 

\subsection{A query formulation of PaLD to add a single datum}
\label{sec:query_pald}

The explicit formula for cohesion \cref{eq:compute_cohension} exposes that three quantities determine its value: set membership in the local foci $U_{x,y}$, the cardinalities $|U_{x,y}|$, and the indicator function values $I(w, x, y)$.  With this in mind, let us consider adding a datum $t$ to $S$ so that 
$T := S \cup \{t\}$ constitutes the extended dataset of interest.  

We then see that 
the cohesion of $w$ to $t$ is given by:
\beq
c_{t,w}^T = \frac{1}{n} \sum_{y \in S} \frac{I(w \in U_{t,y})}{|U_{t, y}^T|} I(w,t,y)
\label{compute_cohension_extended}
\eeq
\noindent
where we employ superscripts to make the use of the extended dataset $T = S \cup \{t\}$ explicit when referring to $U_{t, y}^T$ and $c_{t,w}^T$, as opposed to the previous, implicit notation \cref{eq:compute_cohension} that assumed the reference dataset $S$. Here it is clear that $c_{t,w}^T$  depends only on the relative comparisons that involve the test point $t$.  That is, if we assign the index $(n + 1)$ to $t$, then a reference implementation to compute cohesion to $t$ is given as \cref{alg:cohesionToNew}. 
Here the vector of dissimilarities $D(t, S)$ can be computed in $O(n)$ steps and this computation includes self-cohesion for $t$--i.e., the cohesion of $t$ to $t$.

Conceptually, the cohesion to $t$ always involves $t$ and--since $t$ is fixed--at most $O(n^2)$ comparisons (based on $O(n)$ dissimilarities). 
That is, in \cref{fig_Uxy}, $x=t$ is always present when determining $c_{t,w}^T$, including when determining  $U_{t, y}^T$.

\begin{algorithm}
\caption{CohesionToNew: compute cohesion to a new test point $C_{t, -}^T$}
\label{alg:cohesionToNew}
\begin{algorithmic}
\STATE{Given $D(t,S)=\{ \dd(t,y), \dd(y,t) \}_{y=1}^n$ and $D(S,S)=\{ \dd(x,y)\}_{x,y=1}^n$,} 
\STATE{initialize  $C_{t, -}^T \leftarrow \{ 0\}_{y=1}^n$.}
\FOR{$y = 1$ to $n$}
\FOR{$w = 1$ to $n+1$}
\STATE{$U_{t, y}^T \leftarrow \emptyset$}
\IF{$\dd(w,t) \le \dd(y,t)$ or $\dd(w,y) \le \dd(t,y)$}
\STATE{$U_{t, y}^T \leftarrow U_{t, y}^T \cup \{ w\} $}
\ENDIF
\ENDFOR
\FOR{$w \in U_{x, y}^T$}
\STATE{$C_{t,w}^T \leftarrow C_{t,w}^T + \frac{I(w,t,y)}{|U_{x, y}^T|} $}
\ENDFOR
\ENDFOR
\STATE{ $C_{t, -}^T \leftarrow C_{t, -}^T / n$ }
\RETURN $C_{t, -}^T$
\end{algorithmic}
\end{algorithm}

Asymmetrically, the cohesion of $t$ to $w \neq t$ is given by:
\beq
c_{w,t}^T = \frac{1}{n} \sum_{y \in S} \frac{I(t \in U_{w,y})}{|U_{w, y}^T|}I(t,w,y).  
\label{compute_cohension_from}
\eeq
\noindent
where we emphasize that $U_{w, y}^T$  depends on the extended dataset $T = S \cup \{t\}$. Exact computation of $c_{w,t}^T$ is possible in $O(n^2)$ steps, if  $U_{w, y}^S$ are provided as input.  Since   $U_{w, y}^S$ and  $\{t\}$ are disjoint for distinct $w, y \in S$, either $U_{w, y}^T = U_{w, y}^S$ or $U_{w, y}^T = U_{w, y}^S  \cup \{t\}$.  That is, $|U_{w, y}^T|$ is at most one greater than $|U_{w, y}^S|$ and we have a reference implementation to compute cohesion of $t$ to $w$ in $S$ given as \cref{alg:cohesionToS}.

Intuitively, even though cohesion from $t$ always involves $t$, knowledge of $|U_{w, y}^T|$ is needed to appropriately weight the contributions from $t$ to $c_{w,t}^T$. That is, in \cref{fig_Uxy}, $Z=t$ is always present, but $t$ is not involved in the determining the red points based on the pair $(Y, w=x)$. Even though $t$ is fixed, determination regarding these fourth parties are needed.  On the other hand, if the separation of red points from the grey, green and black points in $U_{w, y}^S$ was determined in advance, then \emph{any} $t$ can be tested against the pair $(Y, w=x)$: first determine if $t \in U_{w, y}^T$ and second, if $t \in U_{w, y}^T$, then determine if $t$ is closer to $w$ than $Y$. 

\begin{algorithm}
\caption{CohesionToS: compute cohesion from a new test point $C_{-, t}^T$}
\label{alg:cohesionToS}
\begin{algorithmic}
\STATE{Given $D(t,S)=\{ \dd(t,y), \dd(y,t) \}_{y=1}^n$,
$D(S,S)=\{ \dd(x,y)\}_{x,y=1}^n$ } 
\STATE{and $U(S,S)=\{ U_{x,y}\}_{x,y=1}^n$, initialize  $C_{-, t}^T \leftarrow \{ 0\}_{y=1}^n$.}
\FOR{$y = 1$ to $n$}
\FOR{$w = 1$ to $n$}
\STATE{$U_{t, y}^T \leftarrow U_{t, y}^S$}
\IF{$\dd(t,w) \le \dd(t, y)$ or $\dd(t,y) \le \dd(w,y)$}
\STATE{$U_{w, y}^T \leftarrow U_{w, y}^T \cup \{ t\} $}
\ENDIF
\ENDFOR
\FOR{$w \in U_{x, y}^T$}
\STATE{$C_{w,t}^T \leftarrow C_{w,t}^T + \frac{I(t,w,y)}{|U_{w, y}^T|} $}
\ENDFOR
\ENDFOR
\STATE{ $C_{-, t}^T \leftarrow C_{-, t}^T / n$ }
\RETURN $C_{-, t}^T$
\end{algorithmic}
\end{algorithm}

The reference implementations given above show that an $O(n^2)$ query algorithm for cohesion is, in general, possible; see \Cref{sec:derive} for additional details.

\begin{theorem}\label{thm:querydetailed}
  Given a reference dataset $S$ of size $n$, a queryable data structure $X_U(S)$ is constructable in $O(n^3)$ time such that for any $t \notin S$ and $T = S \cup \{t\}$, $C_{t,w}^T$ and $C_{w,t}^T$ for all $w \in T$ are computable in $O(n^2)$ steps.
\end{theorem}

Under this query algorithm approach, the memory required is at most $O(n^3)$ to maintain $O(n^2)$ local foci $U_{x,y}$, each of which requires no more that $O(n)$ memory. 

\begin{theorem}\label{thm:querymemory}
  Storing $X_U(S)$ from \cref{thm:querydetailed} requires at most $O(n^3+nd)$ memory, where $d$ is the data dimensionality.
\end{theorem}

In particular, it is possible to iteratively build up the local foci $U_{x,y}$ and dissimilarity data $\dd(x,y)$ needed to compute cohesion. Within the generalized PaLD framework \cite{bfl24}, the contributions to cohesion 
still require at most $O(n^3)$ memory to maintain; see \cref{subsec:considergpald} for discussion.  
For practical applications, sparse representations of the contributions to cohesion can help control the growth of memory, and, if approximations are acceptable (see \cite{bddp22}), memory consumption might be further limited. 
However, detailed consideration of more general settings and the potential for principled approximation techniques are beyond the scope of the present article\footnote{For brief discussion of the complexity challenges of maintaining the full cohesion network, see \cref{sec:challenges}.}.

We now turn to practical considerations for the much simpler situation where a reference data set $S$ is fixed and test points $t$ are considered one at a time. As we illustrate in \Cref{sec:application} below, these assumptions support using labeled $S$ for semi-supervised classification and unlabeled $S$ for online anomaly detection. 

\subsubsection*{Practical considerations and the natural threshold}
For these purposes, we make the standard, mild assumption that:
\beq
 \dd(z, z) <\dd(z, y)  ~{\rm for~all~distinct }~  y, z 
\in T, 
\label{eq:simple}
\eeq
\noindent
and for our implementation for experiments in \cref{sec:speed_up} and \Cref{sec:application}, we take advantage of simplifying assumption that
\beq
\dd(y, z) = \dd(z, y)  ~{\rm for~all }~  y, z 
\in T.
\label{eq:sym}
\eeq
A notable practical consideration is determining the natural threshold $\tau(T, \dd)$ for cohesion—e.g., if the test point $t$ has $w_{t,y} < \tau(T, \dd)$ for all $y \in S$, then $t$  might be considered an outlier. Ideally, we would start from an exact computation of $\tau(T, \dd)$ an introduce any practical approximations from there. After giving an approach to determine $\tau(T, \dd)$ exactly, we describe how to precompute quantities $X$ for the reference set $S$, at an upfront cost of $O(n^3)$, and compute the strong neighborhood of $t$ in $O(n^2)$ steps during during the online stage of the approach.
By caching the cardinalities of local foci for the reference data $S$,
the memory requirements are at most $O(n^2+nd)$, where $d$ is the data dimensionality.

\begin{theorem}\label{thm:querymemorylite}
   The data structure $X_U(S)$ for \cref{thm:querydetailed}, can be replaced by $X_V(S)$ that requires $O(n^2+nd)$ memory, where $d$ is the data dimensionality. Assuming \cref{eq:simple}, the strong threshold $\tau(T, \dd)$ and the strong neighborhood of $t$ can be determined within $O(n^2)$ steps. 
\end{theorem}

The observation that $\tau(S, \dd)$ relates directly to the size of local foci via:
\beq
2n(n-1)\tau(S, \dd) = \sum_{x\in S} \sum_{y\in S, y \neq x} \frac{1}{|U_{x,y}^S|} 
\label{threshold_by_size}
\eeq
\noindent
enables marginal analysis of the impact of adding a single datum $t$. Under \cref{eq:simple}, this leads to an attractive marginal formula for $\tau(T, \dd)$:
\beq
\tau(T, \dd) = \tau(S, \dd) \frac{n-1}{n+1}  + \frac{c_{t,t}^T}{n+1} - \frac{1}{2n(n+1)} \sum_{x\in S} \sum_{y\in S, y \neq x} \frac{I(t \in U_{x,y}^T )}{(|U_{x,y}^S| + 1) |U_{x,y}^S|} 
\label{eq:threshold_marginal}
\eeq
\noindent
in which we can view the right most term 
\beq
\varepsilon(T, \dd) \sdef \frac{1}{2n(n+1)} \sum_{x\in S} \sum_{y\in S, y \neq x} \frac{I(t \in U_{x,y}^T )}{(|U_{x,y}^S| + 1) |U_{x,y}^S|} 
\label{eq:threshold_correction}
\eeq
\noindent
as a correction for those local foci that $t$ is added when extending $S$ to $T$; see \cref{app:details} for derivation details. 

\vspace{.15in}

\noindent {\bf Example.} To get a more ``hands on'' feel for the marginal formula \cref{eq:threshold_marginal}, consider $S = \{z, w\}$. Then,
\beq
2n(n-1)\tau(S, \dd) = \sum_{x\in S} \sum_{y\in S, y \neq x} \frac{1}{|U_{x,y}^S|} = \frac{1}{|U_{z,w}^S|} + \frac{1}{|U_{w,z}^S|} = \frac{1}{2} + \frac{1}{2}
\label{threshold_example}
\eeq
\noindent
where $n = 2$. 

When extending to $T = \{z, w, t\}$, if both $\dd(z, t)$ and $\dd(w, t)$ and are greater than $\dd(z, w)$, then \cref{eq:simple} impiles that the relatively distant test point $t$ will not be added to any local focus.  This implies:
\beqa
2(n+1)n \tau(T, \dd) &=& \sum_{x\in T} \sum_{y\in T, y \neq x} \frac{1}{|U_{x,y}^T|} = 
\frac{1}{|U_{t,w}^T|} + \frac{1}{|U_{t,z}^T|} + \sum_{x\in S} \sum_{y\in T, y \neq x} \frac{1}{|U_{x,y}^T|} \nonumber \\  
&=& \frac{1}{|U_{t,w}^T|} + \frac{1}{|U_{w,t}^T|} + \frac{1}{|U_{w,z}^T|} + \frac{1}{|U_{t,z}^T|} 
+ \sum_{x\in S} \sum_{y\in S, y \neq x} \frac{1}{|U_{x,y}^T|} \nonumber \\  
&=&  2nc_{t,t}^T + 2n(n-1)\tau(S, \dd)
\label{threshold_example_A}
\eeqa
\noindent
where the bottom equality follows from the symmetry of local foci, the fact that local foci do not increase in size for distinct $x, y \in S$, and the previous formula.  Thus, to dividing by $2(n + 1)n$, we have:
\beq
\tau(T, \dd) = \tau(S, \dd)  \frac{n-1}{n+1}  + \frac{c_{t,t}^T}{n+1} 
\label{threshold_example_B}
\eeq
\noindent
with $\varepsilon(T, \dd) = 0$. In this case, $t$ was relatively distant and did not join any existing local foci.  

If $\dd(z, t) < \dd(z, w) < \dd(w, t)$, then:
\beqa
2(n+1)n \tau(T, \dd)
&=& \frac{1}{|U_{t,w}^T|} + \frac{1}{|U_{w,t}^T|} + \frac{1}{|U_{w,z}^T|} + \frac{1}{|U_{t,z}^T|} 
+ \sum_{x\in S} \sum_{y\in S, y \neq x} \frac{1}{|U_{x,y}^T|} \nonumber \\  
&=&  2nc_{t,t}^T + \frac{1}{|U_{z,w}^T|} + \frac{1}{|U_{w,z}^T|} \nonumber \\  
&=&  2nc_{t,t}^T + 2n(n-1)\tau(S, \dd) - \sum_{x\in S} \sum_{y\in S, y \neq x} \frac{1}{(|U_{x,y}^T||U_{x,y}^S|} 
\label{threshold_example_C}
\eeqa
\noindent
where the bottom equality includes a correction term  $\varepsilon(T, \dd) > 0$ since $U_{x,y}^S = U_{y,x}^S$  must be increased when extending $S$ to $T$. Thus, we have a correction term for all (two) local foci for distinct $x, y \in S$ in this case, so that:
\beq
\tau(T, \dd) = \tau(S, \dd)  \frac{n-1}{n+1}  + \frac{c_{t,t}^T}{n+1} - \frac{1}{2n(n+1)} \sum_{x\in S} \sum_{y\in S, y \neq x} \frac{1}{(|U_{x,y}^S|+1)|U_{x,y}^S|}.
\label{threshold_example_D}
\eeq
\noindent For more general $n$, correction terms are introduced based on the indicator $I(t \in U_{x,y}^T )$.

$\square$
\vspace{.15in}

\begin{algorithm}
\caption{PaLDCache: compute data to query strong neighborhoods}
\label{alg:cache}
\begin{algorithmic}
\STATE{ Given $D(S,S)=\{ \dd(x,y)\}_{x,y=1}^n$, initialize  $V \leftarrow \{ 0 \}_{x < y=1}^n$, 
$\tau(S) \leftarrow 0$. }
\FOR{$x = 1$ to $n-1$}
\FOR{$y = x + 1$ to $n$}
\FOR{$w = 1$ to $n$}
\IF{$\dd(w,x) \le \dd(y,x)$ or $\dd(w,y) \le \dd(x,y)$}
\STATE{$V_{x, y} \leftarrow V_{x, y} + 1 $}
\ENDIF
\ENDFOR
\STATE{ $\tau(S) \leftarrow \tau(S) + \frac{2}{V_{x, y}}$ }
\ENDFOR
\ENDFOR
\STATE{ $\tau(S) \leftarrow \frac{\tau(S)}{2n(n-1)}$ }
\RETURN $V$, $\tau(S)$
\end{algorithmic}
\end{algorithm}
        
	We conclude this section with our approach to implementing online PaLD by storing information for queries. We employ a new notation $V_{x,y} = V_{x,y}^S := |U_{x,y}^S|$ and $V_{x,y}^T := |U_{x,y}^T|$, for brevity, as well as our notations cohesion \cref{eq:compute_cohension}, the strong threshold \cref{threshold_by_size}, and the correction term $\varepsilon(T, \dd)$ \cref{eq:threshold_correction}.
    
    After precomputation with the reference data $S$, the strong neighborhood of any new test point $t$ is computed in $O(n^2)$ steps:
    \begin{enumerate}
        \item Cache the sizes of local foci $V_{x,y}$, as when computing $C$, and sum contributions to $\tau(S, \dd)$. This incurs an upfront cost of $O(n^3)$ (\cref{alg:cache}).
        \item Compute the cohesion of $w$ to $t$ in $O(n^2)$ steps (\cref{alg:cohesionToNew}), which does not require cached data as discussed above. 
        \item Compute the cohesion of $t$ to $w \neq t$ in $O(n^2)$ steps using the sizes of the local foci $V_{x,y}$ as input and summing corrections due to the specific local foci whose sizes increase (\cref{alg:queryCohesionToS}).  This modifies \cref{alg:cohesionToS} to use the cached cardinalities from Step 1 and compute $\varepsilon(T, \dd)$. 
        \item Update the strong threshold  via \cref{eq:threshold_marginal}.
        \item Determine the strong neighborhood of $w_{t,y} \geq \tau(T, \dd)$  in $O(n)$ steps by checking if the minimum of $c_{x,y}$   and $c_{y,x}$  (computed in steps 2 and 3) meet or exceed $\tau(T, \dd)$  (computed in step 4).
    \end{enumerate}

\begin{algorithm}
\caption{QueryCohesionToS: compute cohesion from a new test point $C_{-, t}^T$}
\label{alg:queryCohesionToS}
\begin{algorithmic}
\STATE{Given $D(S<t)=\{ \dd(y,t) \}_{y=1}^n$, $D(S<S)=\{ \dd(x,y)\}_{y>x=1}^n$ and } 
\STATE{$V=\{ V_{x,y}\}_{y>x=1}^n$, initialize  $C_{-, t}^T \leftarrow \{ 0\}_{y=1}^n$
$\varepsilon(T, \dd) \leftarrow 0$}
\FOR{$y = 1$ to $n$}
\STATE{$V_{y, y}^T \leftarrow 2$}
\FOR{$w = 1$ to $y-1$}
\STATE{$V_{w, y}^T \leftarrow V_{w, y}^S$}
\IF{$\dd(t,w) \le \dd(y,w)$ or $\dd(t,y) \le \dd(w,y)$}
\STATE{$V_{w, y}^T \leftarrow V_{w, y}^T + 1 $}
\STATE{$\varepsilon(T, \dd) \leftarrow  \varepsilon(T, \dd) + \frac{1}{V_{w, y}^T(V_{w, y}^T -1)}$}
\ENDIF
\ENDFOR
\FOR{$w = 1$ to $n$}
\STATE{$C_{w,t}^T \leftarrow C_{w,t}^T + \frac{I(t,w,y)}{V_{x, y}^T} $}
\ENDFOR
\ENDFOR
\STATE{ $C_{-, t}^T \leftarrow C_{-, t}^T / n$ }
\STATE{ $\varepsilon(T, \dd) \leftarrow \varepsilon(T, \dd) / (n(n+1))$ }
\RETURN $C_{-, t}^T$, $\varepsilon(T, \dd)$
\end{algorithmic}
\end{algorithm}

\subsection{Experimental validation of speed up}
\label{sec:speed_up}

Standard PaLD implementations \cite{dmb25} require $O(n^3)$ comparison steps to analyze a set of $n$ data points. For online PaLD, a second pass of $O(n^2)$ comparisons to reference data is needed each time a new datum is considered. The first, unsupervised, pass discovers communities intrinsic to the reference dataset. The second pass associates a new data point, called a test point, to the original communities in the reference data. When the reference data is labeled, this second pass can be considered a form of semi-supervised learning: the reference labels provide information about test points 
to predict their labels. 

\begin{figure}
    \centering
    \includegraphics[width=0.9\textwidth]{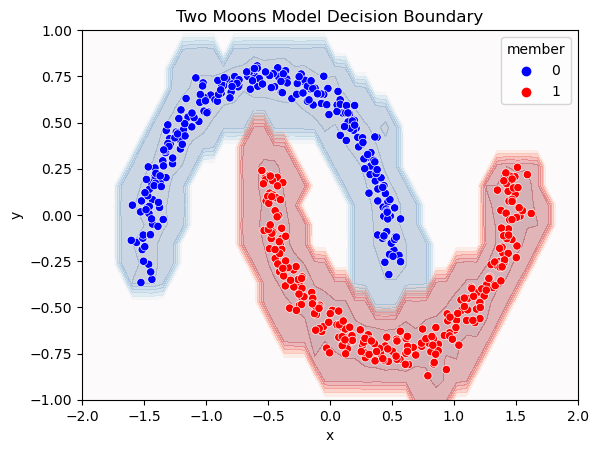}
\caption{Visualization of the decision boundary between two reference classes or clusters.}
\label{fig_decision}
\end{figure}

This type of analysis is similar to the k-nearest neighbors (knn) classification framework where a new test point is assigned the label that represents a best guess based on the labels of its $k$ closest neighbors. 
Ideally, PaLD partitions reference data points into communities such that each community has an intrinsic label. A new test point can then be assigned the same label as the existing community that online PaLD associates it with. 
\cref{fig_decision} provides an example of this semi-supervised use case where online PaLD was used to visualize the decision boundary between two clusters of data.
In \cref{fig_decision}, white areas capture potential test data that might not naturally fall into either class (see \cite{lan22} for comparison and further discussion) and could be considered outliers relative to the labeled reference data.

\subsubsection*{Test Implementation} 
To validate the approach from \cref{sec:query_pald}, we implemented online PaLD in Python. 
As a baseline, we implemented a lightly vectorized version of \cref{alg:paldref}--similar to publicly available R package (https://CRAN.R-project.org/package=pald) \cite{dmb25} with \texttt{numpy} standing in as the vector library.  To build a queryable data structure,  we implemented \cref{alg:cohesionToNew} through \cref{alg:queryCohesionToS}  as a Python interface called \texttt{CohesionDB}.  For example, constructing \texttt{CohesionDB} pays the upfront cost of recording the cardinality of the local foci between every pair of samples found in the reference data (\cref{alg:cache}).  In particular, this supports lazy calculation of all $c_{x,w}^S$ for $x, w \in S$ to get a sense of the communities intrinsic to the reference data $S$ in addition to online computation $c_{x,t}^T$ and $c_{t,x}^T$ for a test point $t$.  Results for lazy computation are reported in \cref{tab:lazy} below.   

\subsubsection*{Results} We used data from \cite{bmm22} and clustering benchmarks \cite{franti2018k,sieranoja2019fast} for speed up validation.
This corresponds to a contrived situation in which a single test point is held out for ``testing'', but keeps our focus on timing. Speed tests were performed on a Dell laptop with an Intel(R) Core(TM) i7-10875H CPU with 64 GB of RAM. 
Our evaluation pipeline consisted of dataset management through the \texttt{DVC} package and performance tabulation using \texttt{mlflow}. 
In the next section, we present initial applications of the approach to health-care data.

We compared our online PaLD implementation to a straightforward Python implementation for datasets of different sizes and captured the processing time along various stages. 
Seven base datasets $T$ of size $n$ were split into a reference set $S$ of size $n-1$ and a single test datum $t$. 
We measured 100 iterations of the following actions: (1) loading the data into an internal structure consisting of nested dictionaries for fast look-up times, (2) querying for the local foci and its cardinality between two points, (3) calculating the cohesion between reference data, (4) calculating the cohesion to a new test point, (5) calculating the cohesion from a new test point, and (6) building the cohesion matrix. 

\begin{table}[bhtp]
\footnotesize
\caption{Mean run times for the baseline PaLD algorithm on $n$ reference data points vs. lazy computation of the cohesion network (averaged  over 100 sample runs).}\label{tab:lazy}
\begin{center}
  \begin{tabular}{|c|c|c|c|c|} \hline
   \multirow{2}{*}{$n$} & \bf Base PaLD (sec) & \multicolumn{3}{c|}{\bf Online PaLD (sec)} \\ \cline{2-5}
   & Network & Loading & Lazy Network & Total \\ \hline
    7 & 2e-3 & 3.9e-4 & 7.6e-4 & 1.1e-3 \\
    15 & 3e-3 & 2e-3 & 4e-3 & 6e-3 \\ 
    239 & 0.97 & 0.64 & 1.3 & 1.9 \\
    499 & 5.9 & 4.6 & 7.8 & 12.4 \\
    787 & 26.6 & 13.8 & 23.5 & 37.3 \\ 
    999 & 64.1 & 28.0 & 45.0 & 72.9 \\ 
    1999 & 612 & 207 & 290 & 497 \\ \hline
  \end{tabular}
\end{center}
\end{table}

Basic runtime comparisons for analyzing a test point $t$ are recorded in \cref{tab:speedup}. For a single test point, online PaLD is faster. In line with complexity analysis, the magnitude of the speedup is roughly scaling with $|S|=n$. 

Comparisons of runtime for computing the cohesion network for the reference data $S$ are given in \cref{tab:lazy}.  For small $S$, the overhead cost of pre-compute and creating the internal data representation (Loading step) leads to slowdown. However, the lazy approach to compute the cohesion network for $S$ may provide speedup opportunities for larger $S$ in terms of total time to compute reference communities.  In any case, if online computation is expected later (which requires the Loading step), it is reasonable to use cached data (from the Loading step) to display and analyze the reference network. The online Lazy Network step to build the cohesion network is trending to be faster than the baseline Network step, so it makes sense to take advantage of data cached during Loading.
%
As $n$ increases toward $10^5$, the Total online runtime may even be preferred but detailed study is beyond our present scope.

Overall, these experimental results exhibit runtimes in line with complexity analysis. In practice, Online PaLD makes online applications with $n=10^5$ accessible with straightforward, if lengthy, training (less than one day) and acceptable query times (less than two minutes). Many interesting health-care data sets fall into this regime and we consider example applications in the next section.    

\section{Applications}
\label{sec:application}

In this section, we present two application areas as illustrations of online PaLD: online anomaly detection and semi-supervised prediction.  Our illustrations provide starting points to help stimulate further study and indicate how online PaLD can speed exploration. 

For online anomaly detection, online PaLD is a novel approach.
At the time of this writing, we are not aware of example applications of PaLD to anomaly detection in print. The examples presented below suggest that while PaLD-based anomaly detection is conceptually similar to $k$-nearest neighbors, it may perform quite differently.

For semi-supervised prediction, we give an example where the number of reference samples is smaller (low $n$) than the dimensionality of each sample (high $d$). Online PaLD is adapted to high $d$, but moderate $n$, applications since (i) PaLD's basis on relative comparisons may help mitigate the curse of dimensionality (cf.\  \cite[Fig.\ 8]{bmm22}) and (ii) its memory requirements collapse the dimensionality of reference data.
That is, the memory needed for online PaLD is dominated by storing the reference data $S$ at $O( n \times d)$ cost for $S$ with sufficiently high dimensionality\footnote{Moreover, reference data can be accessed from disk as opposed to being cached in RAM.}.

Throughout, we employ health-care datasets, which are often not massive and can present challenges that may warrant novel data science approaches. For example, the distribution normal subject data may have different features than anomalous or pathological populations. Within the population of subjects with known pathology, different sub-populations exhibit characteristic features to distinguish modes of pathology  
but have low frequency within the general population.   

\subsection{Online anomaly detection}
\label{sec:anomaly}
A typical finding for diverse anomaly detection benchmarks is that that anomaly detection performance varies significantly by application \cite{han2022adbench}. In practice, experimentation with a variety of anomaly detection algorithms may help tune performance to a specific application. PaLD-based anomaly detection is conceptually similar to $k$-nearest neighbors (knn), including its suitability for data with high dimensionality. 
If PaLD reproduces the performance of knn, then  its value as a new candidate algorithm may be limited. On the other hand, if PaLD behaves differently than knn, then it provides a conceptually similar candidate algorithm.
In these initial experiments, we see that online PaLD may perform quite differently from knn.
Moreover, online PaLD is potentially a good complement to the suite of 14 unsupervised algorithms considered in \cite{han2022adbench}, especially for challengingly anomaly detection benchmarks, which may warrant broader exploration of candidates. 

\subsubsection*{Experimental design}
As a simple test of online PaLD's potential for online anomaly detection, we compare performance to 
$k$-nearest neighbors (knn) for eight anomaly detection benchmarks for health-care data.  Considering benchmarks from ADBench \cite{han2022adbench}, these datasets have sample counts less than 3000 (min: 80, max: 2114), which supports online anomaly detection within seconds\footnote{Note that, without speed up, thoroughly evaluating a benchmark with over two thousand data points is impractical; see, for comparison, \cref{tab:speedup}.}.  Rather than setting specific thresholds for detection, we focus area under curve (AUC) metrics for the receiving operator characteristic (ROC) and precision recall (PR) curves, as in ADBench \cite{han2022adbench}.  

There are choices to derive a test statistic from cohesion (see \cite{lan22}, for comparison, in the context of classification). 
For simplicity, our test statistic is the maximal $w_{x,t}$ \cref{sym_cohesion} value  where $t$ is the datum being tested and $x$ ranges of normal (non-anomaly) reference values $N$\footnote{A related option, for example, would be to consider data labeled as anomalies in a more semi-supervised fashion.}. 
That is, our online test statistic for $t$ is 
\beq
a(t) \sdef \max_{x \in  N \subset S} w_{x,t}^N  
\label{ad_test}
\eeq
\noindent
where $N \subset S$ is the set of non-anomaly values within the reference data $S$ and
lower values $a(t)$ indicate that $t$ is more likely to be an anomaly.
In practice, this means 
that known anomalies have been removed from a normal data to the degree possible.
In terms of parameter-free clustering \cite{bmm22}, a value of $a(t)$ less than the natural threshold would be deemed its own cluster as compared with normal reference values $N$. 
%
%
%
Our knn parameters follow ADBench \cite{han2022adbench} where the default number of near neighbors is 5.  To facilitate comparisons with ADBench, we use its setup to define a new candidate algorithm  based on $a(t)$ \cref{ad_test} so that the details of training and testing splits follow ADBench conventions.    

\subsubsection*{Results}

To indicate the relative difficultly of these anomaly detection problems, we report the metric for the highest performing unsupervised algorithm considered in \cite{han2022adbench}. Our results are summarized in \cref{tab:ad} and suggest that PaLD  distinguishes itself relative to knn and the unsupervised algorithms studied in \cite{han2022adbench}.

We observe differing performance for knn and PaLD with each beating out the other 4 out of 8 times. For knn, we observe perfect scores for Lymphography and WBC, consistent with low numbers of anomalies (6 and 10, respectively) to successfully detect\footnote{For comparison, ROC AUC values of 55.9 and 90.6 for Lymphography and WBC, respectively, appear  in \cite[Table D4]{han2022adbench} for the  knn parameter selections (and randomizations)  made there.}.  For the 4 out of 8 times when PaLD is less performant, the difference is relatively modest. 

Interestingly, PaLD 
\begin{itemize}
    \item outperforms all 14 unsupervised benchmarks from  \cite{han2022adbench} 3 out of 8 times and; 
    \item exhibits a worst-case ROC AUC of over 60\%,
\end{itemize}
 indicating an ability to be at least somewhat informative in a variety of conditions.
Further study of PaLD-based anomaly detection against a broader array of challenging benchmarks would be needed to characterize performance as well as consideration of robustness to our choice of PaLD-based anomaly detection test statistic.

\begin{table}[htbp]
\footnotesize
\caption{Anomaly detection results for 8 healthcare benchmarks for area under curve (AUC) metrics for the receiving operator characteristic (ROC) and precision recall (PR) curves. Cases where the best  performance by an unsupervised algorithm reported in \cite{han2022adbench}  is met or exceeded by PaLD are underlined.}
\label{tab:ad}
\begin{center}
  \begin{tabular}{|l|c|c|c|c|p{15mm}|p{15mm}|} \hline
   \multirow{2}{*}{Dataset} & \multicolumn{2}{c|}{\bf knn} & \multicolumn{2}{c|}{\bf PaLD} & \multicolumn{2}{c|}{\bf Best Unsupervised in \cite{han2022adbench}}\\ \cline{2-7}
   & ROC & PR & ROC & PR & ROC & PR \\ \hline
    breastw & {\bf 99.6} & {\bf 99.1} & 96.9 & 93.3 & 99.7 & 99.4 \\
    cardio &  90.0 & 55.8 & {\bf 95.9} & {\bf 69.4}  & \underline{95.6} & \underline{68.4} \\
    Cardiotocography &  78.1 & 50.8 & {\bf 84.3} & {\bf 63.0}  & \underline{77.8} & \underline{52.6} \\
    Hepetitis &  55.0  & 0.28 & {\bf 63.8} & {\bf 34.5}  & 82.0 & 41.5 \\
    Lymphography & {\bf 1.0} & {\bf 1.0} & 94.2 & 41.2 & 99.8 & 97.6 \\
    Pima &  {\bf 68.5} & {\bf 58.0} & 65.0 & 51.5 & 73.4 & 56.6 \\
    vertebral &  27.7 & 9.1 & {\bf 62.2} & {\bf 18.4}  & \underline{53.2} & \underline{15.2} \\
    WBC & {\bf 1.0} & {\bf 1.0} & 94.8 & 48.0 & 99.5 & 92.3 \\ \hline
  \end{tabular}
\end{center}
\end{table}

\subsection{Semi-supervised prediction}
\label{sec:prediction}
As an application of semi-supervised prediction, we consider a dataset from a recent 
benchmark \cite{xu2023data} for network neuroscience with $n=195$ and $d=4950$.
The Parkinson’s Progression Markers Initiative (PPMI) dataset labels each subject with 4 clinically relevant classes: normal control, scans
without evidence of dopaminergic deficit (SWEDD), prodromal, and Parkinson’s disease (PD). 
These ordinal classes correspond to increasingly severe biomarkers. 
On a practical level, low $n$ is driven by the cost to obtain labels while high $d$ is due to that fact that the matrix of correlations that quantifies a brain network (one per subject) scales quadratically in the number of regions of interest (ROIs) for the brain atlas employed to aggregate voxels from brain scan measurements.  For the present illustration, we consider the Schafer atlas with 100 ROIs, a standard atlas that performs well across many benchmarks in \cite{xu2023data}, including PPMI.  

\subsubsection*{Experimental design}
The present illustration explores classification via online PaLD as compared to knn by holding reference data for evaluation.  Specifically, we use 10 folds to make rough comparisons to the accuracy result reported \cite{xu2023data} possible\footnote{For simplicity, we do not explore parameter tuning, as is done in \cite{xu2023data}.} so that, after randomization\footnote{A stratified shuffle of the data is employed to balance the distribution of labels across folds and the same shuffle is employed across the algorithms evaluated.}, we have 10 test units, each of which trains on 9 out of 10 folds and evaluates accuracy on the remaining 10 percent of the data.
As a result, each labeled correlation matrix in the PPMI benchmark is classified one time. 
An example test unit might have 176 reference data $S$ with 19 labeled correlation matrices $t$ held out to the evaluate accuracy of online prediction.

Following \cite{lan22}, we consider 6 different classification methods derived from the cohesion matrix of the reference data $S$ plus one test point $t$; recall $T = S \cup \{ t\} $.
All six methods measure how strongly $t$ is associated with each of the 4 classes $S_i$ within the labeled data $S$. The class with the strongest association metric is selected to classify the test point $t$: greater values are stronger.

Four methods are based on aggregating information over the neighborhood of $t$, similar to knn.  Two Count metrics consider how many cohesion values exceed the natural threshold $\tau(T, \dd)$ with
\beq
\nu_i(t) \sdef \sum_{x \in S_i} I(c_{x,t} \ge \tau(T, \dd))  
\label{count_metric}
\eeq
\noindent
based on the cohesion to class $S_i$ where $I(c_{x,t} \ge \tau(T, \dd))$ is a 0/1 indicator. A Count metric for the number of strong cohesion values from the each class is defined analogously. Two Sum methods are based on adding the cohesion values that  exceed the natural threshold with
\beq
\sigma_i(t) \sdef \sum_{x \in S_i} I(c_{x,t} \ge \tau(T, \dd)) \cdot c_{x,t}
\label{sum_metric}
\eeq
\noindent
based on the cohesion to class $S_i$; a Sum method for strong cohesion from each class is defined analogously. 

Similar to $a(t)$ defined above \cref{ad_test}, two methods consider maximal cohesion to and from each class. One Max metric is based on cohesion to each class 
\beq
\mu_i(t) \sdef \max_{x \in S_i} c_{x,t};  
\label{max_metric}
\eeq
\noindent
a Max metric for the maximal cohesion from each class is defined analogously. 
See \cite{lan22} for further discussion. 

As in the previous subsection, we do not evaluate the precise reduction in computational time obtained by using online PaLD, which scales with the size of the training data, since we want to take advantage of the greater than $10^2$ speed up.  See \cref{sec:speed_up} for speed up validation.        

\begin{table}[htbp]
\footnotesize
\caption{Mean accuracy across ten folds ($\pm$ one standard deviation) for six cohesion-based classifiers described by \cref{count_metric}, \cref{sum_metric}, and  \cref{max_metric}
 as compared to knn for the PPMI dataset with the Schafer atlas.}
\label{tab:class}
%
\begin{center}
  \begin{tabular}{|c|c|c|c|c|c|c|} \hline
   \multicolumn{3}{|c|}{\bf Cohesion To} & \multicolumn{3}{c|}{\bf Cohesion From} & \multirow{2}{*}{\bf knn}\\ \cline{1-6}
   Count & Sum &  Max & Count & Sum &  Max & \\ \hline
    $0.59 \pm 0.05$ & $0.59 \pm 0.06$ & $0.49 \pm 0.1$ & $0.59 \pm 0.04$ & $0.57 \pm 0.04$ & $0.49 \pm 0.09$ & $0.57 \pm 0.07$ \\ \hline
  \end{tabular}
\end{center}

\end{table}

\subsubsection*{Results}
For this initial experimentation, we observe that the 
two methods using a max cohesion value  underperform knn, but
 the 4 classifiers aggregating over the neighborhood of the test point narrowly outperform knn. 
The 4 strong neighborhood classifiers may even provide lower variation in predictive accuracy as compared knn, but this hypothesis would warrant broader study across benchmarks.

These results are in the general range of accuracy results reported in \cite{xu2023data} where the mean accuracy for PPMI with the Schafer altas lies within $[56.5, 63.2]$ for conventional machine learning methods.
We caution the reader not to place too much weight on these baseline results for the PPMI benchmark  since they do not exploit network structure nor do they systematically address the potential for variation in the measurement process to obtain a corelation matrix for each subject\footnote{There may even be some value in PaLD processing of correlations between ROIs to absorb some of this variation. 
}.
There is much to explore for this benchmark and consideration of PaLD-based classification increases the potential dimensions to explore further. 

The first exploration we pursued was testing the sensitivity of the results to the randomization of the reference data.  That is, if we re-ran the entire experiment, how much variation in the reported mean accuracy should be expected for a given method?
By sampling 10 seeds for randomization, we observed mean accuracies of $0.60 \pm 0.01$ (resp., $0.48 \pm 0.03$) for the count of strong ties to class $i$ (resp., max cohesion to class $i$) method where  $\pm \sigma$ reports the standard deviation of the mean accuracy for each point being tested once, under the random shuffle for a given seed. 
Each sensitivity statistic takes around 220 sec to compute as compared to about 10 hours that would be needed by the original PaLD implementation based on complexity projections.

\section{Derivations and proofs}
\label{sec:derive}

This section provides proofs of our main claims including a derivation of the marginal formula for $\tau(T,\dd)$. 

\subsection{Proofs of main claims}
\label{subsec:proofs}
We begin with a restatement of \cref{thm:query} using the definitions and notations of \Cref{sec:query} that incorporates \cref{thm:querymemorylite}.

\begin{theorem}\label{thm:queryfull}
  Given a reference dataset $S$ of size $n$, the online PaLD algorithm constructs a queryable data structure $X_V(S)$ in $O(n^3)$ time 
  such that for any $t \notin S$ and $T = S \cup \{t\}$ the following quantities are computable in $O(n^2)$ time:
  \begin{itemize}
      \item the cohesion to $t$, $C_{t,w}^T$, and from $t$, $C_{w,t}^T$;
      \item the strong threshold $\tau(T,\dd)$ for the cohesion network $G_T$; and
      \item the strong neighborhood of $t$ within $G_T$.
  \end{itemize}
  with the standard assumption \cref{eq:simple} employed for the latter two quantities. Moreover, storing $X_V(S)$ requires $O(n^2+nd)$ memory, where $d$ is the data dimensionality.
\end{theorem}

As in \cref{alg:cache}, the data structure $X_V(S)$ uses the cardinalities of $V_{x,y} = |U_{x,y}|$ for distinct $x,y \in S$, whereas a more complete data structure $X_U(S)$ encodes $U_{x,y}$ and can be expanded to $X_U^+(S)$ for lazy computation of $c_{x,y}$. 

\subsubsection*{Considerations for a complete data structure} With these distinctions in mind, we treat the more general situation based on $X_U(S)$ before the refinements to reduce required memory via $X_V(S)$.

\vspace{1.5mm}
\noindent{\bf Proof of \cref{thm:querydetailed}:} First, we describe the computation of $C_{t,w}^T$ using stored information. The cohesion of $w$ to $t$, $c_{t,w}^T$ \cref{compute_cohension_extended}, is computable from new (uncached) quantities $U_{t, y}^T$ and $I(w,t,y)$. Once new dissimilarities $D(t, S)$ are computed ($O(n)$ time), each $U_{t, y}^T$ and $I(w,t,y)$ can be determined, as in \cref{alg:cohesionToNew}, with $I(w,t,y)$  a function of $D(t, S)$ and $D(S, S)$ (\ref{IPaLD}). Thus, $\{c_{t,w}^T\}_{w\in T}$ is computable in $O(n^2)$ time and caching $D(S, S)$ may reduce runtime in practice.  

Second, we describe the computation of $C_{w,t}^T$. The cohesion of $t$ to $w \neq t$,  $c_{w,t}^T$ \cref{compute_cohension_from}, is computable from cached $U_{w, y}^S$ and new $I(w,t,y)$. Thus $\{c_{w,y}^T\}_{w\in S}$ is computable in $O(n^2)$ time from $U(S,S)$, $D(t, S)$, and $D(S, S)$, as in \cref{alg:cohesionToS}.  

Finally, we note that all stored information defines
\beq
X_U(S) \sdef (S, D(S,S), U(S,S)).
\label{almost_full_cache}
\eeq
\noindent A $O(n^3)$ reference implementation to compute  $X_U(S)$ as well as  $\tau(S)$  and all support indicators $I(w,x,y)$ is given as \cref{alg:paldcacheref}. 
$\square$\\

\begin{algorithm}
\caption{PaLDCacheFull: reference implementation to compute $X_U^+(S)$}
\label{alg:paldcacheref}
\begin{algorithmic}
\STATE{Given $D(S,S)=\{ \dd(x,y)\}_{x,y=1}^n$,}
\STATE{ initialize  $U(S,S) \leftarrow \{ \emptyset \}_{x< y=1}^n$, $I(S,S,S) := \{ 0 \}_{w,x,y=1}^n$, $\tau(S) \leftarrow 0$.}
\FOR{$x = 1$ to $n-1$}
\FOR{$y = x + 1$ to $n$}
\FOR{$w = 1$ to $n$}
\IF{$\dd(w,x) \le \dd(y,x)$ or $\dd(w,y) \le \dd(x,y)$}
\STATE{$U_{x, y} \leftarrow U_{x, y} \cup \{ w\} $}
\ENDIF
\ENDFOR
\FOR{$w \in U_{x, y}$}
\IF{$\dd(w,x) < \dd(w,y)$}
\STATE{$I(w,x,y) \leftarrow 1 $}
\ENDIF
\IF{$\dd(w,x) = \dd(w,y)$}
\STATE{$I(w,x,y) \leftarrow 1/2$, $I(w,y,x) \leftarrow 1/2 $}
\ENDIF
\IF{$\dd(w,x) > \dd(w,y)$}
\STATE{$I(w,y,x) \leftarrow 1 $}
\ENDIF
\ENDFOR
\STATE{ $\tau(S) \leftarrow \tau(S) + \frac{2}{V_{x, y}}$ }
\ENDFOR
\ENDFOR
\STATE{ $\tau(S) \leftarrow \frac{\tau(S)}{2n(n-1)}$ }
\RETURN $(U(S,S), I(S,S,S), \tau(S)$)
\end{algorithmic}
\end{algorithm}

If a lazy computation is desired, then
\beq
X_U^+(S) \sdef (S, D(S,S), U(S,S), I(S,S,S), \tau(S))
\label{full_cache}
\eeq
\noindent 
provides all the necessary data to compute any cohesion value $c_{x,y}$ in $O(n)$
time via \cref{eq:compute_cohension}. Since $X_U^+(S)$ can be iteratively extended, a cohesion database can be built up gradually and only $I(w, x, y)$ for $w \in U_{x,y}$ need to be stored.

For a practical implementation, the local foci, and if desired, support indicators, can be sparsely represented. We took this approach in \cref{sec:speed_up} for the lazy computation results in \cref{tab:lazy}. If approximations are acceptable, then even less memory may be possible by employing ties; see \cref{sec:challenges} for discussion.

\vspace{1.5mm}
\noindent{\bf Proof of \cref{thm:querymemory}:} The data structure $X_U(S)$ employed for  \cref{thm:querydetailed} requires at most $O(n^3+nd)$ memory since it
maintains:
\begin{enumerate}
    \item  $O(n)$ data of dimension $d$ to store $S$;
    \item $O(n^2)$ distances $D(S,S)$ to help determine $I(w, t, y)$ and $I(t, w, y)$;
    \item $O(n^2)$ local foci, each of which have size $O(n)$; and
    \item $O(1)$ to record $\tau(S)$.  
\end{enumerate}
$\square$

\subsubsection*{Reducing memory requirements}  In \Cref{sec:application}, we gave examples of online anomaly detection and semi-supervised prediction where a single test point $t$ is being compared to $S$.  In this situation, only the quantities itemized in \cref{thm:queryfull} are needed  to reason about how $t$ fits into the (strong) cohesion network.  

For this purpose, we employ
\beq
X_V(S) \sdef (S, D(S,S), V(S,S), \tau(S)),
\label{lite_cache}
\eeq
which includes $\tau(S)$ for thresholding links. Noting that \cref{thm:queryfull}  implies \cref{thm:query} and  \cref{thm:querymemorylite}, we now proceed with the proof of our main theorem.

\vspace{1.5mm}
\noindent{\bf Proof of \cref{thm:queryfull}:} In the proof of \cref{thm:querydetailed}, we showed that the cohesion to and from $t$ are computable from $U(S,S)$, $D(t, S)$, and $D(S, S)$ using the formulas \cref{compute_cohension_extended} and \cref{compute_cohension_from}, respectively. Since these formulae use the cardinalities of local foci, it is sufficient to know $V(S,S)$ to compute cohesion to and from $t$ in $O(n^2)$ time.

If $\tau(S)$ is known, then the correction term \cref{eq:threshold_correction} is computable from $O(n^2)$ summands 
involving $V(S,S)$ so that  $\tau(T)$ follows from \cref{threshold_by_size}.  We derive \cref{threshold_by_size} in \cref{{app:details}}.  Then, the strong neighborhood is computable from $\{c_{t,w}^T\}_{w\in T}$ and $\{c_{w,t}^T\}_{w\in S}$ in $O(n)$ time via \cref{sym_cohesion}.

A $O(n^3)$ implementation to compute  $X_U(V)$ is given by \cref{alg:cache}.  $\square$
 
\subsection{Derivation of the natural threshold formula}
\label{app:details}
Let us now derive a formula for  $\tau(T,\dd)$. First, we note that:
\beq
2n(n-1)\tau(S, \dd) = (n-1) \sum_{x\in S} c_{x,x} = \sum_{x\in S} \sum_{y\in S, y \neq x} \frac{I(x,x,y)}{|U_{x,y}^S|} = \sum_{x\in S} \sum_{y\in S, y \neq x} \frac{1}{|U_{x,y}^S|} 
\label{app_threshold_by_size}
\eeq
since, under \cref{eq:simple}, the left equality follows from the formula for $\tau(S,\dd)$ \cref{threshold}, the middle equality follows the explicit formula for cohesion \cref{eq:compute_cohension} and $I(x,x,y) = 1$ (rightmost equality).  Similarly, we have that:
\beqa
2n(n+1)\tau(T, \dd) &=& n c_{t,t}^T + n \sum_{x\in S} c_{x,x} =  n c_{t,t}^T + \sum_{x\in S} \left(
\frac{I(x,x,t)}{|U_{x,t}^T|} + \sum_{y\in S, y \neq x} \frac{I(x,x,y)}{|U_{x,y}^T|} \right) \nonumber\\ 
&=& n c_{t,t}^T + \sum_{x\in S} \left(
\frac{1}{|U_{x,t}^T|} + \sum_{y\in S, y \neq x} \frac{1}{|U_{x,y}^T|} \right) 
\label{expand_app_threshold_by_size}
\eeqa
by considering $T$ in lieu of $S$ and separating out quantities related to the test point $t$. 
Noting
\beq
\sum_{x\in S} \frac{1}{|U_{x,t}^T|} = \sum_{x\in S} \frac{1}{|U_{t,x}^T|} =  n c_{t,t}^T,
\label{simp_app_threshold_by_size}
\eeq
we have:
\beq
2n(n+1)\tau(T, \dd) = 2n c_{t,t}^T + \sum_{x\in S} \sum_{y\in S, y \neq x} \frac{1}{|U_{x,y}^T|} .
\label{final_app_threshold_by_size}
\eeq
As we discussed during the derivation of \cref{alg:cohesionToS} in \cref{sec:query_pald} above, the set $U_{x,y}^T$ can only differ from $U_{x,y}^S$  by adding the element $t$, so:
\beqa
\sum_{x\in S} \sum_{y\in S, y \neq x} \frac{1}{|U_{x,y}^T|} 
&=& \sum_{x\in S} \sum_{y\in S, y \neq x} \frac{1}{|U_{x,y}^s|} - 
\left( \sum_{x\in S} \sum_{y\in S, y \neq x} \frac{I(t \in U_{x,y}^T )}{(|U_{x,y}^S| + 1) 
|U_{x,y}^S|} \right)
\nonumber\\ 
&=& (n-1) \sum_{x\in S} c_{x,x}^S - 
\left( \sum_{x\in S} \sum_{y\in S, y \neq x} \frac{I(t \in U_{x,y}^T )}{(|U_{x,y}^S| + 1) 
|U_{x,y}^S|} \right)
\label{eq:app_threshold_with_correction}
\eeqa
where $I(t \in U_{x,y}^T )$ is an indicator of set membership and we can view the right most term $\varepsilon(T, \dd)$ as a correction; see \cref{eq:threshold_correction} for comparison.  
 To summarize, we have that:
\beqa
2n(n+1)\tau(T, \dd) &=& 2n c_{t,t}^T + \sum_{x\in S} \sum_{y\in S, y \neq x} \frac{1}{|U_{x,y}^T|} 
\nonumber\\
&=& 2n c_{t,t}^T + (n-1) \sum_{x\in S} c_{x,x}^S - 
\left( \sum_{x\in S} \sum_{y\in S, y \neq x} \frac{I(t \in U_{x,y}^T )}{(|U_{x,y}^S| + 1) 
|U_{x,y}^S|} \right)
\label{app_threshold_by_size_summary}
\eeqa
\noindent
by considering the marginal impact of adding $t$ realized as (i) additional terms in summation formulae and (ii) increases in the size of some local foci.  Upon division by $2n(n+1)$, the attractive marginal formula \cref{eq:threshold_marginal} is obtained. 

\section{Discussion}
\label{sec:discussion}
Online PaLD presents intriguing possibilities for further investigation.  
By making datasets with $n=10^5$ accessible with straightforward training, 
PaLD's potential to be explored prior to bringing in approximations is increased.  
For instance, investigating AD bench \cite{han2022adbench} for datasets with $n < 7000$--including a cross section of computer vision and natural language processing benchmarks \cite{han2022adbench}--could proceed before bringing in approximation to complete the analysis, up to $n=619326$.

Exact PaLD, with its theoretical guarantees, may be attractive to avoid any uncertainty introduced via approximation, especially for situations where reference data is carefully maintained and not too large.
For example, the THINGS initiative (https://things-initiative.org/) developed 1,854 object concepts and curated reference data (such as 26,107 high quality images manually associated to concepts \cite{hebart2019things}). Whereas the original work on PaLD \cite{bmm22}
might provide an interesting unsupervised analysis of these concepts (see \cite{hebart2023things} for comparison), online PaLD is adapted to associate ``test'' concepts (e.g., from user input or some machine interface) to the THINGS data. Comparisons to reference images invite combining the ideas in this article with principled approximation techniques \cite{bddp22} or computer science  approaches \cite{dg24sequential,lan22} to make such comparisons more readily accessible.

As a novel approach to anomaly detection and semi-supervised classification, PaLD is relatively unexplored.
Though not explored here, one intriguing possibility is to combine semi-supervised classification with anomaly detection. Any test point deemed too distant from the training data would simply not be classified. A key question to investigate is how much excluding anomalous test points might increase predictive accuracy. 
As above, it's natural compare with $k$-nearest neighbors, which applies to both 
semi-supervised classification and anomaly detection.
%

\subsection*{Conclusions}
\label{sec:conclusions}
We demonstrated how the partitioned local depth framework can be adapted to online applications.  By analyzing the original, exact algorithm, we limit the complexity of analyzing a test point $t$ relative to reference data $S$.  Our illustrations with health care data show the potential to apply online PaLD to applications where an alternative to the k-nearest neighbor algorithm is desired. 
The online point-of-view we presented complements principled approximation techniques \cite{bddp22} and computer science approaches \cite{dg24sequential,lan22}, inviting future work to hybridize these ideas to study large datasets.
More immediately, online PaLD can help accelerate experimentation across datasets and algorithmic approaches for online applications.  

\section*{Acknowledgments}
We would like to acknowledge helpful discussion of this article with Kenneth Berenhaut and the Metron fellows.

\bibliographystyle{siamplain}
\bibliography{references}

\appendix

\section{Considerations for future work}
\label{sec:future}

For completeness, we discuss considerations to extend online PaLD to generalized PaLD \cite{bfl24} and the challenges inherent in  maintaining a full cohesion network. 

\subsection{An online algorithm for generalized PaLD}
\label{subsec:considergpald}
The generalized partitioned local depth (GPaLD) framework extends PaLD to enable probabilistic consideration of uncertain, variable, and conflicting information \cite{bfl24}.  For instance, GPaLD allows multiple dissimilarities to be combined using probabilistic weights.  This potential to fuse disparate sources of information using the common currency of probability may be relevant to 
semi-supervised classification based on multiple types of predictor variables.
Fortunately, the online algorithm presented is readily adapted to this more general context.

First, absent a notion of distance for GPaLD, we will define what we mean by the online setting in terms of the GPaLD concepts of local relevance and support division. 
Recall that the local relevance of $z$ to the pair $(x, y)$, denoted $R_{x,y,z}$  generalizes the notion of membership within the local focus $z \in U_{x,y}$ to be, in general, a probability such that $R_{x,y,z} = R_{y,x,z}$ and  $R_{x,y,x} = 1$ for all distinct $x, y \in S$ and $z \in S$.  The support division of $z$ with respect the pair $(x, y)$, denoted  $Q_{x,y,z}$, generalizes the indicator $I(z,x,y)$ to be, in general, a probability distribution over $\{ Q_{x,y,z}, Q_{y,x,z} \}$ such that $Q_{x,y,z} = 1 - Q_{x,y,z}$. 
The online setting is concerned with extending the arrays   and for distinct  $x, y \in S$ and $z \in S$, defined over the set $S$, to be defined over the set $T = S \cup \{ t\}$.  

For example, if multiple dissimilarities $\{ {\dd}_i \}_{i=1}^k$ are combined as in \cite[5.1]{bfl24}, then extending this model to the test point $t$ by testing $k$ dissimilarities is consistent with this online setting. In essence, each ${\dd}_i$ gets a (weighted) vote to combine local focus membership and support indicators probabilistically. 

With these preliminaries, the natural threshold derivation is readily adapted. 
Our assumption that $\dd(z, z) < \dd(z, y)$ becomes  $Q_{z,y,z} = 1$  (and hence, $Q_{y,z,z} = 0$ ), which has the social interpretation that $z$ fully supports itself.
The observation that $\tau(S, d)$ relates directly to the size of local foci \cref{threshold_by_size} 
is replaced by the more general measure of the size $V_{x,y}^S := \sum_{w \in S} R_{x,y,w}^S$. The assumption of that $R_{x,y,z} = R_{y,x,z}$ guarantees that $V_{x,y}^S = V_{y,x}^S$  is a function of the set $\{ x, y\}$. 
Likewise, the assumption that $R_{x,y,x} = 1$ \emph{and} $Q_{x,y,x} = 1$  is analogous to $I(x,x,y) = 1$ and we have:
\beq
2n(n-1)\tau(S, \bfR, \bfQ) = \sum_{x\in S} \sum_{y\in S, y \neq x} \frac{1}{V_{x,y}^S} 
\label{threshold_by_size_general}
\eeq
\noindent
Then, marginal analysis of the impact of extending  $\bfR = [R_{x,y,z}]$ and $\bfQ = [Q_{x,y,z}]$  by a single $t$ proceeds as in \cref{app:details} and we update the marginal formula as follows:
\beq
\tau(T, \bfR, \bfQ) = \tau(S, \bfR, \bfQ) \frac{n-1}{n+1}  + \frac{c_{t,t}^T}{n+1} - \varepsilon(T, \bfR, \bfQ)
\label{eq:threshold_marginal_general}
\eeq
\noindent
where $V_{x,y}^T := V_{x,y}^S + R_{x,y,t}$, by definition, and the right most correction term is
\beq
\varepsilon(T, \bfR, \bfQ) \sdef \frac{1}{2n(n+1)} \sum_{x\in S} \sum_{y\in S, y \neq x} \frac{R_{x,y,t}}{(V_{x,y}^S + R_{x,y,t}) V_{x,y}^S} .
\label{eq:threshold_correction_general}
\eeq

\subsection{Challenges for maintaining a full cohesion network}
\label{sec:challenges}
The derivations in \Cref{sec:derive} clarify the challenges in maintaining an exact cohesion network. 
In \cref{app:details}, we see that increases in the size of local foci can be handled systematically in $O(n^2)$ time and, more generally, changes in local relevance are often tractable. 
For the special case of $I(x,x,y)$, a standard assumption determines $I(x,x,y) = 1$ and this makes determination of self-cohesion $c_{x,x}$ and the threshold $\tau(T, \dd)$ tractable.  

However, support indicators $I(w, x,y)$, and support division more generally, are determined by triples of points. Thus, determination of the full cohesion network entails additional computation:
updating the neighborhood around $x \in S$ may require $O(n^2)$ steps.
In essence, to determine $c_{x,y}^T$ for $x \neq y 
\in S$, the division of support due to $t$ must be determined. 
That is, extending the network requires determining $I(t,x,y)$ whenever $t \in U_{x,y}^T$.  In more practical terms, if a new $t$ is inserted deep within $S$, then it will appear in many $U_{x,y}^T$ for $x \neq y 
\in S$ and change many cohesion values.

These considerations suggest lazy computation of $c_{x,y}$ (discussed in \cref{subsec:proofs}) and introducing principled approximation techniques \cite{bddp22}. 
For example, one could consider setting 
$I(t,x,y) := 1 /2$ for $|U_{x,y}^T| > \sqrt{n}$ and analyze how closely this approximates the impact of exact $I(t,x,y)$ on cohesion.   
It would be interesting to compare this proposal to the methods of \cite{bddp22} with $k = \sqrt{n}$.  Such approaches take advantage of the fact that when the local focus $U_{x,y}$ is large, many comparisons may be needed to determine cohesion exactly, but the implied  contributions to cohesion are small. 

\end{document}